\documentclass[times,twocolumn,final,number]{elsarticle}
\biboptions{sort&compress}
\usepackage{medima}
\usepackage{amsmath,amssymb,amsfonts}
\usepackage{algorithmic}
\usepackage{graphicx}
\usepackage{float}
\usepackage{textcomp}
\usepackage{booktabs}
\usepackage{multirow}
\usepackage{array}
\usepackage{makecell}
\usepackage{stfloats}
\usepackage{pifont}

\usepackage{url}
\usepackage[hidelinks]{hyperref}
\usepackage{capt-of}
\usepackage{needspace}
\usepackage{placeins}

\newcommand{\model}{COF}
\newcommand{\registration}{TOPPR}
\newcommand{\SuppSecMethods}{Supplementary Section~1}
\newcommand{\SuppSecSubgroupSummary}{Supplementary Section~2}
\newcommand{\SuppTabSubgroupSummary}{Supplementary Table~1}
\newcommand{\SuppSecSerialPairedFunction}{Supplementary Section~3}
\newcommand{\SuppTabSerialPairedFunction}{Supplementary Table~2}
\newcommand{\SuppSecFocusedStats}{Supplementary Section~4}
\newcommand{\SuppTabFocusedStats}{Supplementary Table~3}
\newcommand{\SuppSecTopologyAudit}{Supplementary Section~5}
\newcommand{\SuppTabTopologyAudit}{Supplementary Table~4}
\newcommand{\SuppSecCrossVisitROI}{Supplementary Section~6}
\newcommand{\SuppFigCrossVisitROI}{Supplementary Fig.~1}
\newcommand{\SuppSecFailureBoundary}{Supplementary Section~6}
\newcommand{\SuppFigFailureBoundary}{Supplementary Fig.~2}
\newcommand{\SuppSecComparatorAudit}{Supplementary Section~7}
\newcommand{\SuppTabComparatorAudit}{Supplementary Table~5}

\journal{Under Review}

\begin{document}
\verso{Haofan Wu et~al.}

\begin{frontmatter}

\title{Chain of Flow: ECG-Conditioned 4D Cardiac Cine Generation from Patient-Specific Anatomical Anchor}

\author[1]{Haofan Wu}
\author[2,3]{Nay Aung}
\author[1]{Theodoros N. Arvanitis}
\author[4]{Joao A. C. Lima}
\author[2,3]{Steffen E. Petersen}
\author[1,2]{Le Zhang\corref{cor1}}
\cortext[cor1]{Corresponding author.}
\ead{l.zhang.16@bham.ac.uk}

\address[1]{School of Engineering, College of Engineering and Physical Sciences, University of
Birmingham, Birmingham, United Kingdom}
\address[2]{William Harvey Research Institute, NIHR Barts Biomedical Research Centre, Queen Mary
University London, London, United Kingdom}
\address[3]{Barts Heart Centre, St Bartholomew's Hospital, Barts Health NHS Trust, London, United Kingdom}
\address[4]{Division of Cardiology, Johns Hopkins University School of Medicine, Baltimore, Maryland, United States}

\begin{abstract}
Cardiac cine magnetic resonance imaging (MRI) is central to functional cardiac assessment, yet a
full current cine sequence may not always be directly available at the point of analysis. We
introduce Chain of Flow (\model{}), an electrocardiography (ECG)-conditioned framework that
combines patient-specific MRI and current ECG for subject-specific 4D cardiac cine generation.
On the UK Biobank dataset, \model{} achieves strong image-level fidelity and downstream
function-oriented performance on a shared same-visit evaluable benchmark. Multi-slice and
multi-resolution analyses indicate stable structural generation quality across the short-axis
stack and heterogeneous acquisition resolutions. Controlled phase-robustness analyses across
resampled input MRI phases further provide same-visit proxy support for patient-specific MRI
plus current ECG when a target MRI phase is not directly observed. A cross-visit route provides
exploratory serial evidence, with the clearest gains in current-facing region-of-interest
readout. Disease-category functional audits, case-level volume-trajectory evidence review
further delineate where the current patient-specific MRI plus ECG formulation remains stable for
anatomy-aware downstream cardiac analysis. Code is available at
\url{https://anonymous.4open.science/r/COF-paper-release-C88B}.
\end{abstract}


\end{frontmatter}
\phantomsection\label{frontmatter:medima_article_info}

\section{Introduction}
\label{sec:introduction}

Cardiovascular diseases remain the leading cause of morbidity and mortality worldwide \cite{roth2020global}. 
Accurate evaluation of cardiac anatomy and function is essential for diagnosis, risk
stratification, and treatment planning \cite{kramer2008standardized,potter2018assessment}.
High-precision quantification of chamber volumes, myocardial deformation, and regional function
provides clinicians with valuable information for supporting clinical decision-making
\cite{american2002standardized}.
Cardiac cine magnetic resonance imaging (MRI) is uniquely capable of capturing detailed anatomy
together with time-resolved motion, making it central to contemporary functional cardiac imaging
\cite{gray2018patient,petersen2016reference}. This role builds on longstanding MRI
motion-imaging formulations that linked cardiac phase and anatomy within one acquisition
paradigm \cite{axel1989mr}. Large reference cohorts further anchor its role in quantitative
morphology and function assessment \cite{real2023magnetic}. It also provides an important
imaging foundation for patient-specific anatomy-aware cardiac analysis and modelling
\cite{corral2020digital,coorey2022health}. Yet broader MRI-plus-ECG cine-support motivations
remain. A current target temporal point may not always be directly observed by MRI even when a
patient-specific MRI is already available in hand \cite{xia2021recovering}, and serial
reassessment may later rely on prior MRI when repeat cine is delayed or unavailable
\cite{betemariam2025barriers}. These scenarios motivate methods that combine patient-specific
MRI with current temporal information. The present paper evaluates this formulation first on a
controlled same-visit benchmark and then examines missing-target-phase and cross-visit serial
settings as additional scoped analyses.

The electrocardiogram (ECG) offers an attractive complementary source of subject-specific
temporal dynamics because it is inexpensive, ubiquitous, and continuously measurable
\cite{attia2019artificial}. However, ECG signals capture electrical activity without explicit
anatomical structure, making direct recovery of full 4D cardiac anatomy using ECG as the only
input severely ill-posed \cite{rudy1999noninvasive}. In practice, a more clinically grounded
formulation is to use ECG as a temporal driver while conditioning generation on patient-specific
MRI. This shifts the task to generating ECG-consistent cine dynamics from patient-specific MRI
rather than from ECG alone. In the present paper, the main experiments focus on a controlled
same-visit benchmark in which the MRI input is available and fixed. Two additional analyses
examine same-visit proxy support when the target cine phase is not directly observed and
exploratory cross-visit serial use when prior MRI is available from an earlier visit. In all
three settings, patient-specific MRI supplies subject-specific structure and current ECG
supplies the contemporaneous temporal signal used to update that structure.

In this work, we present Chain of Flow (\model{}), an ECG-conditioned framework for
subject-specific 4D cardiac volume generation from patient-specific MRI and current ECG. The
core idea is to separate patient-specific structural context from dynamic motion modelling: a
registration-guided anatomical teacher provides anatomy-aware motion supervision, and an
ECG-conditioned flow model learns to generate temporally coherent cardiac motion consistent with
the input ECG. This formulation preserves clinically meaningful structure while explicitly
acknowledging that patient-specific MRI is part of the inference input. Within this scoped
setting, we study whether ECG-guided generation can produce an analyzable 4D cardiac cine
representation from usable MRI input and current ECG.

\begin{figure*}[t]
    \centering
    \includegraphics[width=2\columnwidth]{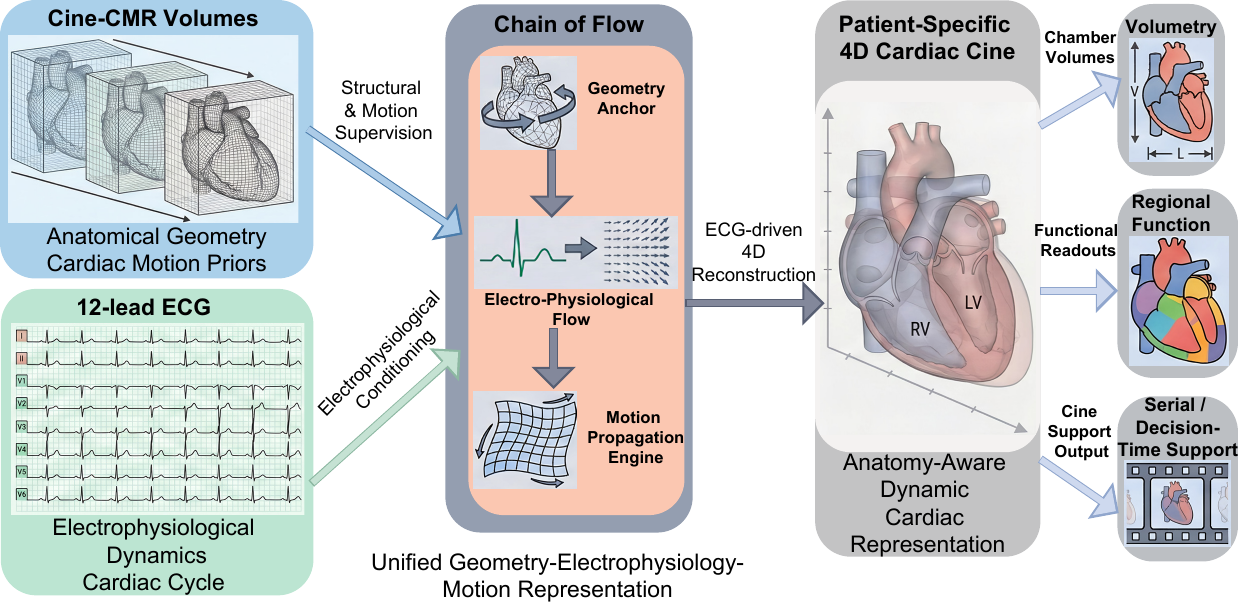}
\caption{Overview of \model{} for ECG-guided 4D cardiac volume generation from patient-specific
MRI and current ECG. The framework integrates CMR volumes and 12-lead ECG during training to
learn a unified representation of cardiac anatomy, electrophysiology, and motion dynamics.
Cine-CMR provides anatomical geometry and motion supervision, while ECG supplies
subject-specific electrophysiological dynamics over a single cardiac cycle. These multimodal
signals are fused within the proposed Chain of Flow, which couples geometry encoding,
electro-physiological flow, and motion propagation to form an ECG-conditioned dynamic
representation. At inference, the learned model takes patient-specific MRI together with current
ECG to generate a subject-specific 4D cine sequence with individualised cardiac motion. The
resulting representation supports downstream analyses including chamber volumetry, regional
functional assessment, and anatomy-aware dynamic cardiac analysis.}
    \label{fig:toc}
\end{figure*}

To study this scoped setting, we evaluate \model{} on paired ECG--CMR data using a layered
protocol: a controlled same-visit benchmark on a shared evaluable benchmark, additional analyses
of same-subject anchor variation and cross-visit serial use, and robustness/failure-boundary
audits. Clinically, these analyses cover same-visit ECG-conditioned cine generation from
patient-specific MRI, controlled proxy support when the current target cine phase is not
directly observed, and exploratory cross-visit serial support when only prior MRI is available.
An overview of the framework is shown in Fig.~\ref{fig:toc}. The main contributions of this work
are as follows:

\begin{itemize}
    \item We propose \textbf{\model{}}, an ECG-conditioned framework that combines
        patient-specific MRI and current ECG for subject-specific 4D cardiac cine generation.
    
    \item We introduce a \textbf{registration-guided anatomical teacher with ECG-conditioned
        flow matching} to separate anatomy from dynamics and generate physiologically plausible
        cardiac motion.
    
    \item We perform \textbf{protocol-consistent validation} on a shared same-visit benchmark
        together with anchor-robustness analyses, exploratory serial analyses, and
        robustness/boundary audits. These evaluations test whether the generated cines can
        support downstream cardiac function analysis from fixed patient-specific MRI while also
        characterizing controlled missing-target-phase proxy support and exploratory cross-visit
        serial support.
\end{itemize}

\section{RELATED WORK}

\subsection{Clinical Motivations for Anatomy-Anchored Cine Generation}
Cardiac cine MRI remains central to chamber volumetry and regional function assessment
\cite{bellenger2000comparison,lang2015recommendations}. It is also a core modality for broader
functional cardiac evaluation \cite{gray2018patient}. Related time-resolved CMR families further
extend haemodynamic and flow readouts in clinical practice
\cite{bissell20234d,demirkiran2022clinical}. Two motivating use readings help frame the present
work. First, current-facing support may be needed when a target temporal point is not directly
observed by MRI even though a subject-specific reference MRI is already available and ECG can
still be acquired rapidly and inexpensively \cite{attia2019artificial,xia2021recovering}.
Second, serial reassessment may later rely on a prior MRI-derived anchor when repeat cine is
delayed or unavailable in routine care \cite{betemariam2025barriers}. Existing work on
incomplete cardiac MRI has mainly focused on recovering missing or unusable image content from
imaging inputs themselves \cite{xia2021recovering}. That setting is related but distinct: it
does not address the clinically motivated problem of combining reference MRI with current ECG to
support current-facing cine interpretation at a missing target phase. The longer-horizon
literature on anatomy-aware cardiac modeling helps motivate richer dynamic representations, but
it does not remove the need for cine support from the MRI and ECG inputs that are actually
available at decision time \cite{trayanova2012computational,xie2022physics}. These scenarios
motivate the present ECG-conditioned cine formulation, which we evaluate first on a controlled
same-visit anatomy-anchored benchmark.

\subsection{Generative Models for Cardiac Structure and Motion}
Generative modelling has recently gained traction in cardiac imaging for reconstructing
anatomical structure, synthesising cine sequences, and improving image quality
\cite{chartsias2018factorised,yi2019generative}. Early medical image synthesis work helped
establish this direction for anatomy-aware reconstruction and enhancement \cite{nie2018medical}.
Variational autoencoders, diffusion models, and flow-based generative frameworks have been
applied to tasks such as cine reconstruction, motion estimation, and cross-modality synthesis
\cite{meng2023deepmesh,qin2018joint}. Cross-modality synthesis pipelines also rely on paired or
weakly paired structural correspondence across imaging domains
\cite{chartsias2017multimodal,zhu2017unpaired}.
Recent medical video generation has also expanded to echocardiography and privacy-preserving
cardiac video synthesis \cite{reynaud2023feature,reynaud2024echonet}. Video diffusion backbones
provide a broader temporal generation template beyond cardiac imaging \cite{ho2022video}. More
recent cardiac-specific generation work extends flow matching to 3D+t shape completion and
generation, but remains conditioned on imaging inputs rather than ECG at inference
\cite{ma2025cardiacflow}.
These methods learn latent representations of cardiac geometry and temporal deformation,
improving robustness to noise and enabling realistic 3D or 4D cardiac generation
\cite{meng2023deepmesh,qin2018joint}. However, existing cardiac generation pipelines generally
remain conditioned on explicit imaging inputs at inference, such as cine CMR \cite{qin2018joint}
or paired cross-modality images, which provide dense spatial supervision unavailable from ECG
alone. Cross-modal synthesis variants still assume shared spatial structure or visually aligned
volumetric data, an assumption that breaks down when ECG is the available temporal signal.
Purely perceptual video realism can also blur boundaries, distort chamber geometry, or weaken
myocardial-thickness consistency in cardiac settings \cite{blau2018perception,li2024echopulse}.
Echocardiography motion-estimation studies likewise show that temporal realism alone does not
guarantee anatomy-faithful deformation readout \cite{evain2022motion}. Incomplete-MRI imputation
methods address degraded imaging inputs \cite{xia2021recovering}, but they do not solve
anatomy-anchored, ECG-driven dynamic cine reconstruction from reference MRI plus ECG in the
missing-target-timepoint or serial-support settings. This leaves a method gap for cardiac
generation that must simultaneously preserve anatomy, maintain temporally coherent motion, and
admit ECG-based conditioning at inference.

\subsection{ECG-Based and Multimodal Cardiac Modeling}

The 12-lead electrocardiogram is one of the most accessible and widely used non-invasive tools
for cardiac assessment \cite{hong2020opportunities,somani2021deep}.
Prior studies have leveraged ECG to estimate global functional parameters such as heart-rate
variability and conduction delays \cite{attia2019prospective}. Others classify arrhythmias and
ischaemic abnormalities \cite{hannun2019cardiologist,ansari2023deep}. Recent machine-learning
reviews further emphasise ECG's diagnostic reach across acute ischaemia and occlusion-level
triage \cite{al2023machine}. In the context of structural heart disease, several studies have
leveraged ECG signals to infer myocardial infarction (MI) characteristics such as infarct
location \cite{engelen1999value} and transmural extent \cite{akbar2024acute}. Others infer
electrical activation abnormalities from the QRS complex or ST--T patterns
\cite{selvester1985selvester}. While these approaches demonstrate the diagnostic value of ECG,
their outputs are inherently low-dimensional and do not capture the full geometric or dynamic
complexity of the heart. Classical inverse-ECG formulations instead reconstruct epicardial
potentials and activation maps from body-surface recordings \cite{oster1997noninvasive}.
Subsequent sparsity-based and wavelet-promoted regularization strategies improve numerical
stability but remain tied to the same ill-posed inverse setting
\cite{ghosh2009application,cluitmans2018wavelet}. Spatiotemporal regularization solvers follow
the same trajectory while still focusing on electrical activation rather than anatomy-aware cine
generation \cite{messnarz2004new}. However, the inverse problem is severely ill-posed: distinct
activation patterns can produce similar ECG observations, and small measurement noise can lead
to large variations in the inferred solutions \cite{macleod1998recent,rudy1999noninvasive}. As a
result, inverse ECG frameworks focus on electrical activity alone, without reconstructing
cardiac geometry or mechanical motion \cite{pereira2020electrocardiographic}.

Recent multimodal work has begun to couple ECG with imaging through joint latent modelling or
cross-modal information transfer \cite{beetz2022multi,turgut2025unlocking}. Related studies also
combine ECG with CMR \cite{nielsen2013computing} or CT \cite{ramanathan2004noninvasive}.
Electroanatomic-mapping frameworks likewise target localised physiological properties rather
than subject-specific cine generation \cite{dhamala2020embedding,li2024toward}. Related cardiac
model-personalization strategies optimize over graph- or atlas-based priors rather than cine
synthesis \cite{dhamala2019bayesian}. Nevertheless, these approaches still depend on explicit
imaging input at inference \cite{li2024solving} or remain limited to task-specific,
low-dimensional outputs \cite{arevalo2016arrhythmia}. They may align, predict, or summarise
structure--function relationships, but they do not generate current-facing 4D cine from
reference MRI plus current ECG. More specifically, they do not cover the two motivating use
readings considered here: same-cycle proxy support when the target cine phase is not directly
observed by MRI, and cross-visit exploratory support when a usable MRI anchor is available but
updated temporal information must come from ECG. This leaves a gap for ECG-conditioned,
reference-anchored, temporally coherent cardiac cine generation that remains suitable for
downstream functional analysis.

\begin{figure*}[!t]
    \centering
    \includegraphics[width=2\columnwidth]{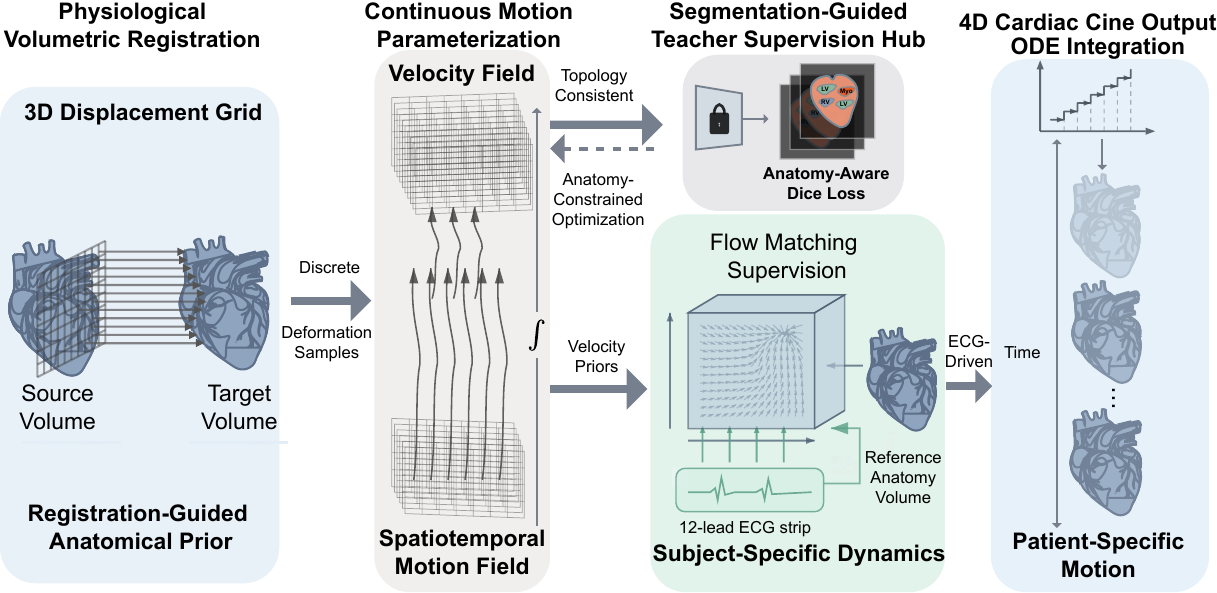}
\caption{Algorithmic pipeline of registration-guided motion modelling and ECG-conditioned cine
generation. The pipeline starts with a volumetric registration teacher that estimates 3D
displacement fields between source and target CMR volumes at different cardiac phases. Discrete
deformation samples obtained from registration are parameterised as a continuous spatiotemporal
velocity field. A segmentation-guided teacher supervision hub constrains the registration using
anatomy-aware Dice loss. The resulting subject-specific motion dynamics are used as supervision
for flow matching, linking ECG signals to physiological motion priors. Finally, the learned
velocity field is integrated over time via ODE-based integration to produce a 4D cine sequence
with patient-specific cardiac motion.}
    \label{fig:framework}
\end{figure*}

\section{METHODOLOGY}

\subsection{Problem Formulation and Inputs/Outputs}
Let $\mathcal{J}_i \in \mathbb{R}^{D \times T \times H \times W}$ denote the preprocessed 4D
cine volume of subject $i$, and let $\mathbf{e}_i \in \mathbb{R}^{12 \times L}$ denote one
temporally aligned 12-lead ECG cycle. Rather than predicting anatomy from ECG alone, \model{}
assumes access to a patient-specific reference anatomical volume
\begin{equation}
\mathbf{R}_i = \mathcal{J}_i(:,0,:,:)\in \mathbb{R}^{D \times H \times W},
\end{equation}
which corresponds to frame~0 of the preprocessed cine sequence used by the current
implementation. In this paper, frame~0 is therefore the canonical reference anchor, rather than
a claim that every subject is perfectly aligned to a universal end-diastolic identity after
resampling. The task is to learn a generator
\begin{equation}
G_{\theta} : (\mathbf{R}_i, \mathbf{e}_i) \mapsto \hat{\mathcal{J}}_i,
\end{equation}
where $\hat{\mathcal{J}}_i \in \mathbb{R}^{D \times T \times H \times W}$ is a temporally
coherent 4D cine sequence whose anatomy remains anchored to $\mathbf{R}_i$ while its motion is
conditioned by the ECG. Equivalently, the model predicts a time-varying deformation field
$\hat{\mathcal{U}}_i=\{\hat{\mathbf{u}}_{i,t}\}_{t=1}^{T}$ and warps the reference anatomy to
obtain the full sequence. In the controlled benchmark used here, $\mathbf{R}_i$ is instantiated
by frame~0 of the same cine study; clinically, the broader role of $\mathbf{R}_i$ is a usable
MRI-derived anatomical anchor available at inference.

Cohort routing, ECG-cycle extraction, and cine preprocessing are specified in
Sec.~\ref{sec:datasets_preprocessing}.

\subsection{Stage 1: Registration-Guided Anatomical Prior}
The first stage provides anatomy-aware motion supervision through pairwise registration. For a
reference frame $\mathbf{R}_i$ and a target phase volume
$\mathbf{V}_{i,t}=\mathcal{J}_i(:,t,:,:)$, we estimate a displacement field
$\mathbf{u}_{i,t}(\mathbf{x})\in\mathbb{R}^{3}$ such that
\begin{equation}
\hat{\mathbf{V}}_{i,t}(\mathbf{x})=\mathcal{W}(\mathbf{R}_i,\mathbf{u}_{i,t})(\mathbf{x})=\mathbf{R}_i(\mathbf{x}+\mathbf{u}_{i,t}(\mathbf{x})).
\end{equation}
The registration backbone is a 3D VoxelMorph-style network \cite{balakrishnan2019voxelmorph}
that operates on the preprocessed low-resolution canonical volumes and is trained on
reference--target pairs sampled from the same subject across cardiac phases. Although this stage
is used as a motion teacher for Stage~2, it is not a one-off classical registration
pre-processing step. Instead, it is a reusable differentiable module that produces consistent
low-resolution flow supervision for every frame and exposes anatomy features that are reused
downstream. In the ablation table we refer to this stage by the module token \registration{}.

Stage~1 is trained with a composite objective. First, an intensity-based reconstruction loss
measures similarity between the warped reference volume and the target phase:
\begin{equation}
\mathcal{L}_{\mathrm{rec}} = 1-\mathrm{NCC}\!\left(\hat{\mathbf{V}}_{i,t}, \mathbf{V}_{i,t}\right),
\end{equation}
where $\mathrm{NCC}$ denotes local normalised cross-correlation. Second, we impose segmentation
consistency using a frozen nnU-Net teacher $\mathcal{S}_{\phi}$ trained on manually annotated
CMR data \cite{isensee2021nnu}. The warped prediction and target volume are mapped into the
nnU-Net prediction space, the teacher branch produces hard anatomical labels, and the largest
connected component of each foreground class is retained to suppress noisy fragments. The
student branch keeps soft probabilities to preserve gradient flow. The resulting multi-class
Dice supervision is
\begin{equation}
\mathcal{L}_{\mathrm{seg}} = 1 - \frac{1}{C-1}\sum_{c=1}^{C-1}
\frac{2\sum_{\mathbf{x}} P_{\mathrm{stud},c}(\mathbf{x})Y_{\mathrm{teach},c}(\mathbf{x})}
{\sum_{\mathbf{x}} P_{\mathrm{stud},c}^{2}(\mathbf{x}) + \sum_{\mathbf{x}} Y_{\mathrm{teach},c}^{2}(\mathbf{x}) + \varepsilon},
\end{equation}
where $C$ is the number of classes including background. The Stage~1 objective is
\begin{equation}
\mathcal{L}_{\mathrm{Stage1}} = \lambda_{\mathrm{rec}}\mathcal{L}_{\mathrm{rec}} + \lambda_{\mathrm{seg}}\mathcal{L}_{\mathrm{seg}}.
\end{equation}
Applying the learned registration from the fixed reference frame to every target phase yields a
subject-specific sequence of anatomy-consistent teacher flows,
\begin{equation}
\mathcal{U}_i = \{\mathbf{u}_{i,1},\dots,\mathbf{u}_{i,T}\},
\end{equation}
which becomes the motion prior for Stage~2.

\subsection{Stage 2: ECG-Conditioned Dynamic Flow Matching}
Stage~2 learns a continuous ECG-conditioned motion generator from the Stage~1 teacher flows.
Each teacher displacement field is packed together with a confidence channel into a
low-resolution flow tensor $x_{0}\in\mathbb{R}^{C\times T\times H_f\times W_f}$, where $C=4D_f$
and $D_f$ is the low-resolution depth produced by the registration teacher. The model then
learns a conditional velocity field that transports Gaussian noise toward this teacher
trajectory, following recent continuous transport views of generative modeling
\cite{lee2024improving}.

The conditioning signal combines anatomy and ECG dynamics. A frozen STMEM encoder
\cite{na2024guiding} maps the input ECG cycle to a 768-dimensional embedding
$\mathbf{c}_{\mathrm{ecg}}$. A reference-anatomy branch reuses the Stage~1 VoxelMorph teacher
\cite{balakrishnan2019voxelmorph} to encode $\mathbf{R}_i$ into spatial features. We
additionally learn a 128-dimensional embedding of a monotone phase vector
$\boldsymbol{\phi}\in[0,1]^T$ to capture temporal reparameterisation. The resulting
896-dimensional dynamic vector is combined with the spatial reference features inside a 3D U-Net
flow-matching network. In the ablation study, removing the reference-anatomy anchoring branch is
denoted by ``w/o REA''.

Given a teacher flow tensor $x_{0}$, we sample a Gaussian endpoint
$x_{1}\sim\mathcal{N}(0,\mathbf{I})$ and an interpolation time $\tau\sim\mathcal{U}(0,1)$, and
form the standard flow-matching state
\begin{equation}
x_{\tau} = (1-\tau)x_{0} + \tau x_{1}, \qquad
\mathbf{v}^{*} = x_{1}-x_{0}.
\end{equation}
The network receives $x_{\tau}$, the aligned anatomy features, the scalar time image $\tau$, and
the dynamic conditioning vector, and predicts a velocity field
$\mathbf{v}_{\theta}(x_{\tau},\tau,\mathbf{c})$. Training minimises the mean-squared error
between the predicted and target velocities,
\begin{equation}
\mathcal{L}_{\mathrm{FM}}
=
\mathbb{E}_{\tau,x_{0},x_{1}}
\left[
\left\|
\mathbf{v}_{\theta}(x_{\tau},\tau,\mathbf{c})
- \mathbf{v}^{*}
\right\|_{2}^{2}
\right].
\end{equation}
This formulation ties the ECG-conditioned dynamics to a subject-specific motion prior rather
than directly regressing anatomy from the ECG.

\subsection{Inference: From Continuous Flow to 4D Cine}
At inference, the model receives only the reference anatomy $\mathbf{R}_i$ and one ECG cycle
$\mathbf{e}_i$. The same anatomy and ECG encoders provide spatial and temporal conditions, after
which the flow-matching model integrates the learned velocity field from noise to the
teacher-flow domain. The current implementation uses explicit Euler integration with 20 steps,
consistent with standard neural ordinary differential equation (ODE) discretization practice
\cite{chen2018neural}:
\begin{equation}
x^{(k+1)} = x^{(k)} + \Delta\tau \, \mathbf{v}_{\theta}\!\left(x^{(k)},\tau_k,\mathbf{c}\right), \qquad \tau_k: 1 \rightarrow 0.
\end{equation}
Under this reverse-time schedule, $\Delta\tau < 0$ at each Euler step.
The resulting flow pack is unpacked into a sequence of displacement fields
$\hat{\mathcal{U}}_i=\{\hat{\mathbf{u}}_{i,t}\}_{t=1}^{T}$, upsampled to the full volume
resolution, and applied to the reference anatomy:
\begin{equation}
\hat{\mathcal{J}}_i(\mathbf{x}, t)
=
\mathcal{W}\!\left(\mathbf{R}_i,\hat{\mathbf{u}}_{i,t}\right)(\mathbf{x}).
\end{equation}
Because the same reference volume is warped across all time points, the generated cine sequence
preserves subject-specific anatomy while allowing ECG-conditioned motion to evolve over the full
cardiac cycle.

\subsection{Implementation Details and Reproducibility Hooks}
Both stages are implemented in PyTorch. Stage~1 uses the segmentation-aware canonical
configuration for 100 epochs (batch size 8, learning rate $2\times10^{-4}$) with multi-scale
reconstruction and Dice-based teacher supervision. Stage~2 uses the same canonical subject split
for 1{,}200 epochs (batch size 4, learning rate $2\times10^{-4}$, drops at epochs 800 and
1{,}000). A single release path registry standardises caches, checkpoints, teacher models,
evaluation resources, cine reconstruction, geometry restoration, independent assessment, and
downstream metrics; extended registration-teacher, ECG-conditioning, subject-disjoint evaluator,
and image-metric details are summarised in \SuppSecMethods{}.

\Needspace{5\baselineskip}
\section{EXPERIMENTS AND RESULTS}

\subsection{Set-up}
\subsubsection{Datasets and preprocessing}\normalcolor
\label{sec:datasets_preprocessing}

We evaluate \model{} on paired ECG and short-axis cine CMR data derived from a 10,000-case
paired UK Biobank entrance cohort \cite{petersen2016uk}. Table~\ref{tab:protocol_transparency}
summarises the cohort routing used throughout this manuscript. All benchmark statistics are
reported on the final shared evaluable benchmark obtained after protocol-specific filtering
together with common export and evaluator screening.

\begin{table*}[t]
\centering
\caption{Protocol transparency summary for the cohorts used in this manuscript. The table
records the auditable routing from model development to the shared evaluable benchmark,
evaluator training cohort, and cross-visit serial cohort.}
\setlength{\tabcolsep}{3pt}
\renewcommand{\arraystretch}{1.08}
\begin{tabular}{p{42mm}p{72mm}p{10mm}p{34mm}}
\toprule
Protocol stage & Cohort definition & $N$ & Use in study \\
\midrule
Generation entrance cohort & Paired UKB ECG--CMR cohort & 10{,}000 & generation route \\
Shared benchmark & Final same-visit benchmark & 426 & main results \\
Evaluator source & Annotated 4D source cohort & 600 & assessor source \\
Overlap removed & Subject overlap removed & 9 & independence audit \\
Evaluator-training cohort & Subject-disjoint assessor cohort & 591 & evaluator route \\
Cross-visit serial cohort & Dual-visit serial cohort & 62 & serial analysis \\
\bottomrule
\end{tabular}
\label{tab:protocol_transparency}
\end{table*}

The generation entrance cohort defines the paired ECG--CMR source pool, the shared benchmark
supports the main same-visit tables, the 591-case subject-disjoint evaluator-training cohort is
used only to train the independent automatic assessor, and the 62-case dual-visit cohort is
reserved for exploratory serial analysis. Within the final benchmark cohort, the mean age is
$65.31 \pm 7.75$ years, the mean heart rate is $61.51 \pm 11.63$ bpm, the sex distribution is
44.6\% female and 55.4\% male, and the ECG-status breakdown is 21.6\% ECG-normal, 77.9\%
ECG-abnormal, and 0.5\% ECG-unknown.

For ECG preprocessing, resting 12-lead ECG signals are sampled at 500~Hz and converted to
millivolts using the acquisition metadata. R-peaks are detected using NeuroKit2, and a
representative single-cycle segment between successive R peaks is cropped to provide one
subject-specific cardiac cycle of electrical activity. For CMR preprocessing, the raw UK
Biobank short-axis cine input remains a 50-frame cardiac sequence, while the current
implementation maps it onto a temporally resampled canonical grid at spatial size
$128\times128$. The first available cine frame is used as the canonical reference anatomical
anchor throughout training and inference; in the current data flow this is frame~0, but it
should not be read as meaning that every case is strictly identical to end-diastole after
canonical resampling. Each training example pairs one reference anatomical volume with one
temporally aligned ECG cycle.

To obtain anatomy-aware supervision during Stage~1, manually annotated short-axis images were
used under expert supervision to train the frozen nnU-Net teacher \cite{isensee2021nnu}. For
evaluator-dependent analysis, we train a separate 3D SegResNet-based automatic assessor
\cite{cardoso2022monai} on the subject-disjoint evaluator route summarised in
Table~\ref{tab:protocol_transparency}. This keeps the final anatomy/function readout separate
from the nnU-Net supervision route at both the subject and model-family levels, while the
image-level metrics are computed first on the same shared evaluable benchmark before any
evaluator-dependent analysis. Final evaluator-dependent results are then computed only on that
shared evaluable benchmark by rebuilding subject-level 4D labels and passing them through this
independent SegResNet assessor.

\subsubsection{Evaluation Metrics}\normalcolor
We separate the evaluation into two groups to make the independence scope explicit.
\textbf{Image-level fidelity metrics.}
In the main paper, we report DINOv2-based Fr\'echet temporal distance (FVD) and
DINOv2-based Fr\'echet image distance (FID) \cite{oquab2023dinov2}, together with peak
signal-to-noise ratio (PSNR) and structural similarity (SSIM), as the primary image bundle.
These metrics are computed directly from the real and generated cine volumes and do not use
automatic segmentation outputs. In the canonical evaluation pipeline, the FID score uses
subject-level DINOv2 features obtained by averaging slice--frame embeddings within each 3D+t
volume, while the FVD score uses a subject-level temporal embedding built from the ordered
mid-ventricular slice sequence with the same DINOv2 backbone.
\textbf{Independent-evaluator anatomical agreement metrics.}
To assess geometric fidelity across the full heart, both the real and generated cine volumes are
reconstructed into subject-level 4D sequences and processed by the independent automatic
anatomical assessor trained on the subject-disjoint cohort described above. Slice-wise and
volume-wise Dice score, intersection-over-union (IoU), and 95th-percentile Hausdorff distance
(HD95) are then computed between the rebuilt labels from the real and generated cine volumes.
\textbf{Independent-evaluator functional agreement metrics.}
We further evaluate the suitability of \model{} for downstream functional cardiac analysis by
deriving end-diastolic volume (EDV), end-systolic volume (ESV), ejection fraction (EF), stroke
volume (SV), left-ventricular volume--time curves, and phase landmarks from the same
independently generated labels. Table~\ref{tab:downstream_seg} emphasises a five-metric function
bundle: ESV MAE, EF MAE, SV MAE, volume-curve root-mean-square error (RMSE), and ES phase error.

\begin{table*}[!t]
\centering
\caption{Primary image-level comparison on the shared evaluable benchmark. Lower is better for
FVD/FID. FVD and FID scores are reported as mean $\pm$ spread from 50 resampled
evaluations on aligned 426-subject DINOv2 feature bundles \cite{oquab2023dinov2}, while PSNR
and SSIM are reported as mean $\pm$ standard deviation across subjects. Here, FVD and FID denote
DINOv2-based Fr\'echet-style temporal and image distribution distances, rather than standard
Inflated 3D or Inception-feature implementations.}
\setlength{\tabcolsep}{5.0pt}
\renewcommand{\arraystretch}{1.12}
\begin{tabular*}{\textwidth}{@{}>{\raggedright\arraybackslash}p{24mm}@{\extracolsep{\fill}}rrrr@{}}
\toprule
Method & \multicolumn{1}{c}{FVD$\!\downarrow$} & \multicolumn{1}{c}{FID$\!\downarrow$} & \multicolumn{1}{c}{PSNR$\!\uparrow$} & \multicolumn{1}{c}{SSIM$\!\uparrow$} \\
\midrule
\textbf{\model{}} & \textbf{12.92$\pm$0.5} & \textbf{4.54$\pm$0.1} & \textbf{28.74$\pm$0.85} & 0.985$\pm$0.004 \\
COF-w/Diff. & 31.60$\pm$0.7 & 8.82$\pm$0.2 & 26.86$\pm$0.77 & 0.978$\pm$0.005 \\
X-Dyna & 19.38$\pm$0.6 & 7.28$\pm$0.1 & 27.12$\pm$0.84 & 0.984$\pm$0.004 \\
MoFA-Video & 47.24$\pm$1.5 & 19.48$\pm$0.4 & 27.01$\pm$0.80 & 0.982$\pm$0.004 \\
EchoDiffusion & 98.00$\pm$1.8 & 66.48$\pm$0.9 & 28.53$\pm$0.73 & \textbf{0.986$\pm$0.004} \\
ECHOPulse & 608.24$\pm$7.7 & 437.35$\pm$3.0 & 11.87$\pm$0.63 & 0.361$\pm$0.066 \\
ConsistI2V & 229.62$\pm$4.8 & 294.57$\pm$1.7 & 13.09$\pm$0.61 & 0.541$\pm$0.054 \\
LFDM & 536.76$\pm$6.1 & 457.68$\pm$4.1 & 11.41$\pm$0.26 & 0.146$\pm$0.033 \\
EchoNet-Syn. & 509.92$\pm$4.0 & 471.97$\pm$2.8 & 15.97$\pm$0.61 & 0.732$\pm$0.036 \\
\bottomrule
\end{tabular*}
\label{tab:sota_compare_compact}
\end{table*}

\subsection{ECG-conditioned 4D Cardiac Cine Reconstruction}
We first test whether the controlled reference-anatomy + ECG formulation can recover analyzable
4D cine volumes on the 426-subject shared evaluable benchmark defined in
Table~\ref{tab:protocol_transparency}. This same-visit benchmark evaluates full-sequence
generation before any segmentation-based evaluator is applied and judges performance using FVD,
FID, PSNR, and SSIM. Table~\ref{tab:sota_compare_compact} reports image-level fidelity under
this protocol. These linked comparison tables form the paper's formal same-visit comparative
evaluation on the shared evaluable benchmark across the retained cardiac and video-generation
baselines. \SuppSecComparatorAudit{} and \SuppTabComparatorAudit{} summarise why several
additional cardiac/ECG-related comparator families are not yet eligible for evaluator-closed
main-table comparison under their current accessible routes. This includes ECGFlowCMR
\cite{fang2026ecgflowcmr} and CardiacFlow \cite{ma2025cardiacflow}. The same audit also covers
the single-frame-to-dynamic Sequence-Aware Diffusion Model (SADM) line \cite{wang2023sadm}. In
the linked comparison tables, COF-w/Diff.\ denotes COF-w/ Diffusion, and EchoNet-Syn.\ denotes
EchoNet-Synthetic.

\begin{figure}
    \centering
    \includegraphics[width=0.96\columnwidth]{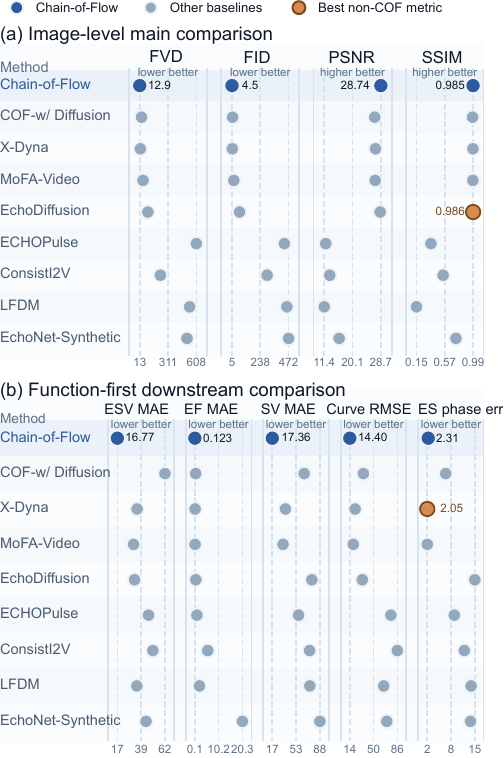}
\caption{(a) Image-level tradeoff view aligned with the main comparison in
Table~\ref{tab:sota_compare_compact} after direction-aware normalisation. Larger area indicates
a better overall image-level balance across the four reported metrics. (b) Function-first
comparison aligned with the main comparison in Table~\ref{tab:downstream_seg} after the same
direction-aware normalisation to summarise the downstream ranking.}
    \label{fig:radar}
\end{figure}

Within this image-level comparison bundle, \model{} leads the table on FVD (12.92), FID (4.54),
and PSNR (28.74) while remaining in the top SSIM band (0.985). EchoDiffusion is marginally
highest on SSIM (0.986), but it trails \model{} substantially on the core perceptual-temporal
pair FVD/FID. The added FID/FVD dispersion summaries preserve the same coarse ranking picture.
The case-aggregated PSNR/SSIM spreads also remain tight across the shared evaluable cohort. The
image-side results are therefore reported descriptively, whereas paired inference is reserved
for the function-facing surface where case-level evaluator outputs are available for all
pre-specified comparisons. Fig.~\ref{fig:radar}(a) visualises these four image-level tradeoffs
in one direction-aware view.

\begin{figure}[!b]
    \centering
    \includegraphics[width=\columnwidth]{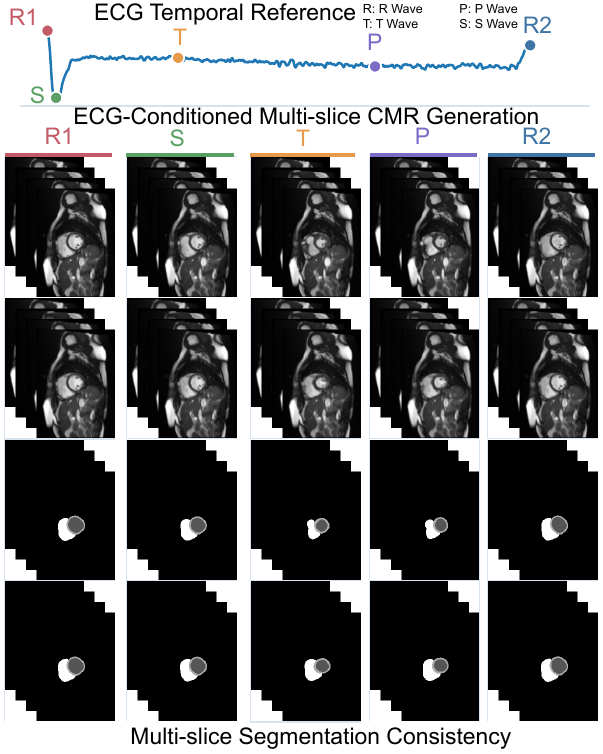}
\caption{ECG-conditioned 4D cardiac cine generation results. The model conditions on a full
12-lead electrocardiogram, while the top trace shows one displayed lead from the aligned cardiac
cycle for visual reference; the R-wave and T-wave markers indicate the temporal landmarks used
to orient the displayed ED/ES phases within the cine sequence. The figure also shows the
short-axis slice-stack context and the generated multi-slice cine examples across selected
cardiac phases. For each phase, the four rows show the reference CMR slices, the generated
slices, the reference labels from the real cine sequence, and the labels obtained from the
generated slices through the automatic anatomical evaluator.}
    \label{fig:result}
\end{figure}

\subsection{Downstream Functional Utility for Cardiac Analysis}
Having established image-level generation quality on the same shared evaluable benchmark, we
next test whether the generated cine remains useful for downstream cardiac analysis.
Table~\ref{tab:downstream_seg} evaluates the same benchmark after rebuilding subject-level 4D
sequences and processing both real and generated cine volumes through the same subject-disjoint
independent automatic evaluator. The primary downstream bundle is deliberately function-first:
we report ESV MAE, EF MAE, SV MAE, volume-curve RMSE, and ES phase error. To strengthen this
function-facing surface without expanding to a full all-method significance matrix, we
additionally summarise 95\% bootstrap confidence intervals for COF together with paired
difference support against COF-w/Diff., X-Dyna, and MoFA-Video in \SuppSecFocusedStats{} and
\SuppTabFocusedStats{} using exact paired sign tests with Holm correction across the five
pre-specified primary function metrics for each comparator family.

\begin{table*}[!t]
\centering
\caption{Primary downstream function comparison under the independent automatic evaluator. Lower
is better for all five reported metrics.}
\setlength{\tabcolsep}{5.0pt}
\renewcommand{\arraystretch}{1.06}
\begin{tabular*}{\textwidth}{@{}>{\raggedright\arraybackslash}p{24mm}@{\extracolsep{\fill}}rrrrr@{}}
\toprule
Method & \multicolumn{1}{c}{ESV MAE$\!\downarrow$} & \multicolumn{1}{c}{EF MAE$\!\downarrow$} & \multicolumn{1}{c}{SV MAE$\!\downarrow$} & \multicolumn{1}{c}{Vol.\ curve RMSE$\!\downarrow$} & \multicolumn{1}{c}{ES phase err.$\!\downarrow$} \\
\midrule
\textbf{\model{}} & \textbf{16.77} & \textbf{0.123} & \textbf{17.36} & \textbf{14.40} & 2.31 \\
COF-w/Diff. & 62.07 & 0.476 & 64.97 & 34.92 & 6.95 \\
X-Dyna & 35.57 & 0.236 & 37.04 & 22.47 & \textbf{2.05} \\
MoFA-Video & 32.18 & 0.217 & 33.32 & 19.67 & 2.07 \\
EchoDiffusion & 33.17 & 0.531 & 76.42 & 33.38 & 14.70 \\
ECHOPulse & 46.39 & 0.997 & 56.58 & 76.66 & 9.24 \\
ConsistI2V & 50.54 & 5.606 & 73.11 & 86.40 & 11.91 \\
LFDM & 35.27 & 2.087 & 73.46 & 65.77 & 13.72 \\
EchoNet-Syn. & 44.03 & 20.320 & 88.44 & 70.37 & 13.55 \\
\bottomrule
\end{tabular*}
\label{tab:downstream_seg}
\end{table*}

Within this bundle, \model{} is the strongest overall method, achieving the best ESV MAE, EF
MAE, SV MAE, and volume-curve RMSE. The only column it does not win is ES phase error, where
X-Dyna (2.05) and MoFA-Video (2.07) are slightly lower than \model{} (2.31). The focused paired
support in \SuppSecFocusedStats{} sharpens this reading: COF remains lower than COF-w/Diff.,
X-Dyna, and MoFA-Video on ESV MAE, EF MAE, SV MAE, and volume-curve RMSE, whereas ES phase error
is the one exception, remaining clearly better than COF-w/Diff., not lower than X-Dyna under the
paired test, and does not show a COF advantage over MoFA-Video. Across the primary downstream
bundle, the generated cine preserves strong volumetric and functional agreement, with ES phase
error as the only readout that does not support a uniform COF advantage. Together, these results
show that the generated cine remains usable for volumetry, phase-aware functional readout, and
related cardiac analysis once a usable anatomical anchor is available.

\Needspace{8\baselineskip}
\subsection{Fine-Grained Image-Quality Validation}
\label{sec:fine_grained_image_quality}
\subsubsection{Slice-wise analysis}
\begin{figure*}
    \centering
    \includegraphics[width=0.92\textwidth]{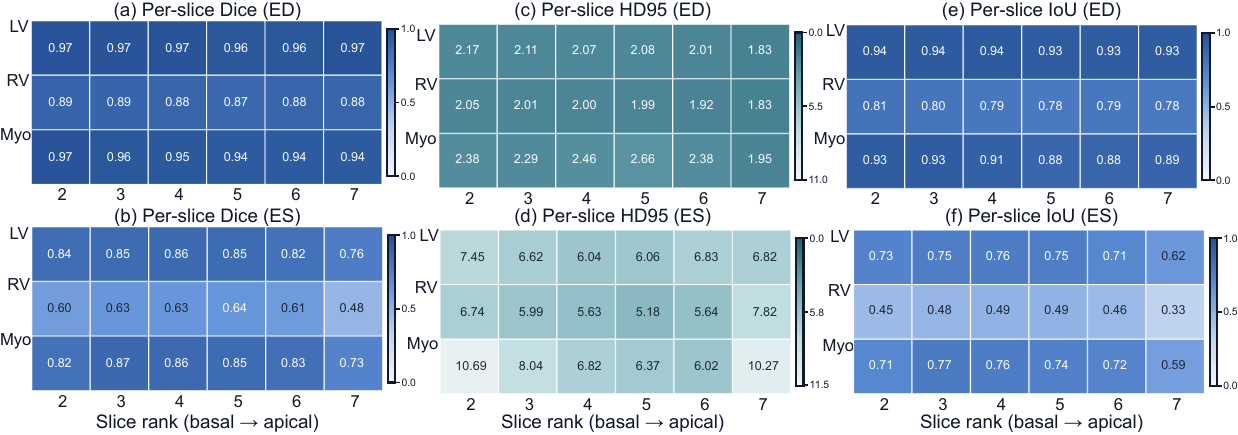}
\caption{Per-slice anatomical agreement across cardiac phases. Slice-wise metrics are reported
at end-diastole and end-systole after processing both real and generated cine with the
independent automatic anatomical assessor. Slices are ordered from basal to apical (ranks 2--7)
and evaluated separately for the left ventricle, right ventricle, and myocardium. Dice and IoU
heatmaps (a, b, e, f) use a normalised 0--1 scale, whereas HD95 heatmaps (c, d) report 2D
per-slice surface distances computed with in-plane spacing. Each cell is the subject-mean value
for the corresponding structure and slice rank.}
    \label{fig:slice_hotmap}
\end{figure*}

\normalcolor
With the main benchmark now established through Table~\ref{tab:sota_compare_compact},
Table~\ref{tab:downstream_seg}, and Fig.~\ref{fig:radar}, we next use more fine-grained analyses
to test whether the generated cine remains stable at the image-quality and boundary-readout
level. We begin with a slice-wise evaluation over the full short-axis stack of generated 4D
volumes. For each subject, we evaluate two key cardiac phases—end-diastole (ED) and end-systole
(ES)—and compute anatomical agreement metrics between the rebuilt labels from the real and
generated cine sequences for the left ventricular cavity, right ventricular cavity, and
myocardium under the independent automatic evaluator. We adopt a data-driven anatomical grouping
strategy based on myocardium coverage. We compute the myocardium area for each short-axis slice
and identify the slice with maximal myocardium area as the anatomical centre. Slices with
negligible myocardium presence are excluded by thresholding their area to at least $25\%$ of the
peak value. These slices are then ordered along the basal–apical axis and indexed by their
relative slice rank. For each valid slice rank, structure, and phase (ED/ES), we compute Dice,
IoU, and HD95. Dice and IoU are evaluated on a per-slice basis, while HD95 is computed in 2D
using the original in-plane spatial resolution from the CMR header. Metrics are aggregated
across subjects to obtain mean performance profiles as a function of slice rank, which are
visualised using slice-wise heatmaps for ED and ES (Fig.~\ref{fig:slice_hotmap}). These ED/ES
profiles show that boundary agreement remains well preserved across LV, RV, and myocardium from
basal to apical slices, with support from all three readouts: Dice, IoU, and HD95.

\Needspace{8\baselineskip}
\subsubsection{Resolution-wise robustness}
\begin{figure}[!t]
    \centering
    \includegraphics[width=0.94\columnwidth]{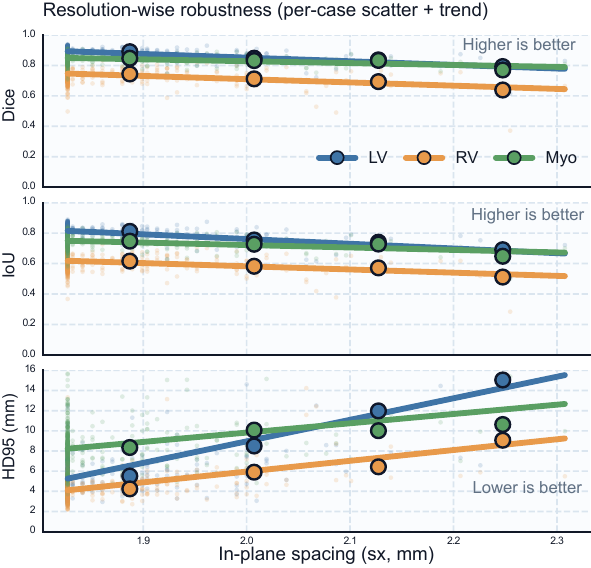}
\caption{Resolution-wise robustness. Dice (top), IoU (middle), and HD95 (bottom) are plotted
against the in-plane pixel spacing $s_x$ (mm); points denote individual subjects, solid lines
linear trend fits, and outlined markers spacing-bin means for LV, RV, and myocardium.}
    \label{fig:resolution_robustness}
\end{figure}

\normalcolor
To assess whether \model{} remains robust under heterogeneous acquisition resolutions, we
conduct a resolution-wise analysis using the in-plane pixel spacing $s_x$ extracted from the CMR
header of each subject. For every case, we compare the rebuilt label volumes from the generated
and real cine sequences for LV, RV, and Myo under the independent automatic evaluator, and
compute Dice, IoU, and HD95. Concretely, for each subject we first compute \emph{per-frame}
metrics on the 3D volumes at all $T$ timepoints and then average them over time to obtain
subject-level scores. HD95 is computed on 3D surfaces using the original voxel spacing $(s_x,
s_y, s_z)$. We additionally record the per-case $s_x$ and retain all subject-level metrics for
subsequent visualisation. We then partition the cohort into $N_{\text{bins}}$ non-overlapping
spacing intervals that uniformly cover the full range of observed $s_x$ values. Within each bin,
we aggregate the subject-level metrics for each structure and report the mean and standard
deviation. These statistics are visualised using grouped bar plots with error bars. Beyond
binned summaries, we further visualise resolution trends using per-case scatter plots of metric
versus $s_x$ together with structure-wise linear trend lines; the mean performance of each
spacing bin is overlaid as larger markers (Fig.~\ref{fig:resolution_robustness}). Across the
full spacing range, \model{} exhibits stable ED/ES boundary agreement and no systematic
degradation in Dice, IoU, or HD95 with increasing pixel spacing.

\Needspace{4\baselineskip}
\subsection{Anchored and Serial Support Analyses}
We next examine two additional clinically motivated settings beyond the main same-visit
benchmark: a controlled same-visit anchor-shift proxy for missing-target-phase support and an
exploratory cross-visit serial route for region-of-interest (ROI) readout. One analysis follows
the resampled same-subject anchor grid, and the other follows the prior-MRI + current-ECG serial
setting.

\subsubsection{Same-Subject Reference-Anchor Phase Analysis}
We evaluate anchor robustness by fixing the subject and visit, shifting MR and ECG together
across the resampled 11-anchor grid, and comparing all canonical anchor frames on the shared
evaluable benchmark. Table~\ref{tab:reference_anchor_all11} reports FVD, motion correlation,
motion SSIM, macro Dice, LV Dice, and EF MAE across these anchors and shows broadly stable
performance, indicating that no single anchor uniquely drives the selected image/anatomy
results. Residual variation is still visible in function-oriented columns such as EF MAE, but
the overall pattern remains anchor-robust across the shared same-visit protocol. This
same-subject anchor-shift analysis therefore provides controlled proxy support for
missing-target-phase use.

\begin{table}[!t]
\centering
\caption{Reference-anchor phase analysis on the shared evaluable benchmark. Subject and visit
are fixed while MR and ECG are shifted together across 11 canonical anchor frames; frame~0 is
the default inference anchor. Columns report FVD, motion correlation (Corr.), motion SSIM, macro
Dice, LV Dice, and EF MAE.}
\setlength{\tabcolsep}{2.6pt}
\renewcommand{\arraystretch}{1.03}
\begin{tabular*}{\columnwidth}{@{\extracolsep{\fill}}crrrrrr@{}}
\toprule
Frame & \multicolumn{1}{c}{FVD} & \multicolumn{1}{c}{Corr.} & \multicolumn{1}{c}{SSIM} & \multicolumn{1}{c}{Macro} & \multicolumn{1}{c}{LV} & \multicolumn{1}{c}{EF} \\
\midrule
0 & 15.10 & 0.52 & 0.91 & 0.77 & 0.84 & 0.31 \\
1 & 13.40 & 0.49 & 0.91 & 0.75 & 0.83 & 0.48 \\
2 & 13.29 & 0.45 & 0.91 & 0.75 & 0.82 & 0.54 \\
3 & 13.99 & 0.47 & 0.91 & 0.77 & 0.84 & 0.34 \\
4 & 13.92 & 0.47 & 0.91 & 0.77 & 0.84 & 0.34 \\
5 & 14.24 & 0.43 & 0.91 & 0.75 & 0.82 & 0.60 \\
6 & 12.86 & 0.44 & 0.91 & 0.75 & 0.82 & 1.15 \\
7 & 12.40 & 0.48 & 0.91 & 0.76 & 0.83 & 0.60 \\
8 & 11.30 & 0.48 & 0.91 & 0.78 & 0.84 & 0.77 \\
9 & 12.33 & 0.48 & 0.91 & 0.78 & 0.85 & 0.35 \\
10 & 14.98 & 0.50 & 0.91 & 0.77 & 0.84 & 0.30 \\
\bottomrule
\end{tabular*}
\label{tab:reference_anchor_all11}
\end{table}

\subsubsection{Cross-Visit Serial ROI and Prior-Trajectory Analysis}
We finally examine a 62-subject cross-visit serial route to test whether registration improves
current-facing ROI agreement when prior MRI is paired with current-visit ECG.
Table~\ref{tab:serial_roi_support} summarizes the current-direction serial analysis over
all 62 currently available same-subject dual-visit subjects for whom prior-visit MRI,
current-visit MRI, and visit-aligned current-visit ECG are all available, as summarised in
Table~\ref{tab:protocol_transparency}. Prior acts as the naive carry-forward reference,
Current+ECG provides a same-visit ROI upper-bound context, and the main serial comparison is
between Prior+ECG and Reg.prior+ECG. On this cohort, Reg.prior+ECG substantially improves the
current-facing ROI readout over Prior+ECG, raising LV Dice from 0.571 to 0.731, Myo Dice from
0.545 to 0.640, and Macro Dice from 0.452 to 0.607. By contrast, the unregistered Prior+ECG
branch does not improve ROI alignment over the carry-forward prior, indicating that ECG
conditioning alone is insufficient for cross-visit spatial update without registration. The
prior-trajectory function readouts remain mixed: prior-curve correlation rises from 0.854 to
0.908, but Prior+ECG retains the lower EF/SV drift means (0.080/10.592 versus 0.083/11.234). The
paired subject-level serial summary mirrors this split. On the ROI side, Reg.prior+ECG is better
in 58/62 subjects for LV Dice and in 58/62 subjects for Myo Dice. It is better in 61/62 subjects
for Macro Dice, indicating that the current-facing ROI gains are not driven by only a handful of
outlier subjects. On the prior-trajectory side, the paired counts remain mixed: Prior+ECG
retains higher prior-curve correlation in 34/62 subjects, lower prior EF drift in 36/62, and
lower prior SV drift in 33/62; \SuppSecSerialPairedFunction{} and \SuppTabSerialPairedFunction{}
therefore continue to support only an exploratory serial reading. These serial results support
cross-visit current-facing cardiac ROI readout when prior-to-current registration brings the
cardiac region into closer correspondence, while the prior-trajectory readouts remain mixed.

\begin{table*}[!t]
\centering
\caption{Cross-visit serial support on the 62-subject same-subject dual-visit cohort. Prior
denotes the carry-forward baseline, Prior+ECG the unregistered serial branch, Reg.prior+ECG the
registered serial update, and Current+ECG the same-visit upper-bound context. The first three
columns report current-facing ROI Dice, followed by prior-trajectory curve correlation and EF/SV
drift. The last three columns report prior-trajectory stability rather than current-visit
functional accuracy.}
\setlength{\tabcolsep}{4.2pt}
\renewcommand{\arraystretch}{0.98}
\begin{tabular*}{\textwidth}{@{\extracolsep{\fill}}lrrrrrr@{}}
\toprule
Method & \multicolumn{1}{c}{LV Dice} & \multicolumn{1}{c}{Myo Dice} & \multicolumn{1}{c}{Macro Dice} & \multicolumn{1}{c}{Prior curve Corr.} & \multicolumn{1}{c}{Prior EF drift} & \multicolumn{1}{c}{Prior SV drift} \\
\midrule
Prior & 0.583 & 0.552 & 0.464 & 1.000 & 0.000 & 0.000 \\
Prior+ECG & 0.571 & 0.545 & 0.452 & 0.854 & 0.080 & 10.592 \\
Reg.prior+ECG & 0.731 & 0.640 & 0.607 & 0.908 & 0.083 & 11.234 \\
Current+ECG & 0.894 & 0.858 & 0.832 & 0.950 & 0.067 & 10.346 \\
\bottomrule
\end{tabular*}
\label{tab:serial_roi_support}
\end{table*}

\begin{figure*}[!t]
    \centering
    \includegraphics[width=\textwidth]{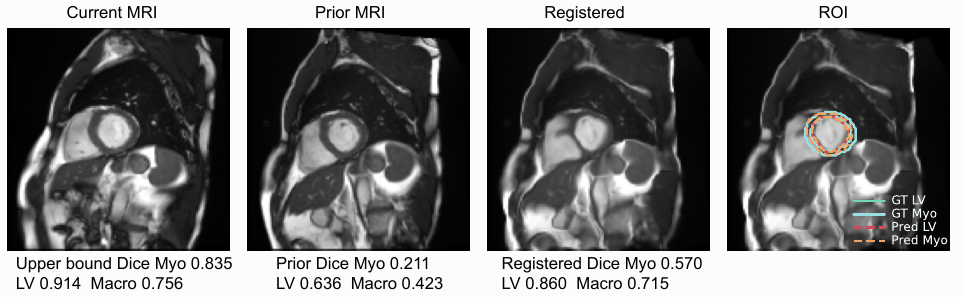}
\caption{Representative case-level visual support for the cross-visit serial ROI analysis.
Panels show the current MRI, the prior MRI, the registered update, and the ROI contour
comparison. In the ROI panel, solid cyan/light-blue contours denote the ground-truth
LV/myocardium, and dashed red/orange contours denote the registered prediction.}
    \label{fig:serial_roi_support_case}
\end{figure*}

Figure~\ref{fig:serial_roi_support_case} shows a representative ROI-facing serial case. The
displayed Dice rows and contours show that the registered update yields a visibly tighter
cardiac contour match than the prior baseline while remaining below the same-visit upper bound,
with a complementary second serial case shown in \SuppSecCrossVisitROI{} and
\SuppFigCrossVisitROI{}.

Table~\ref{tab:serial_registration_qc} explains why the ROI is the most readable cross-visit
target under the current registration route: the cardiac ROI improves much more strongly than
the non-ROI, while the remaining registered absolute error still lies predominantly outside the
ROI.

\begin{figure*}[!t]
    \centering
    \includegraphics[width=\textwidth]{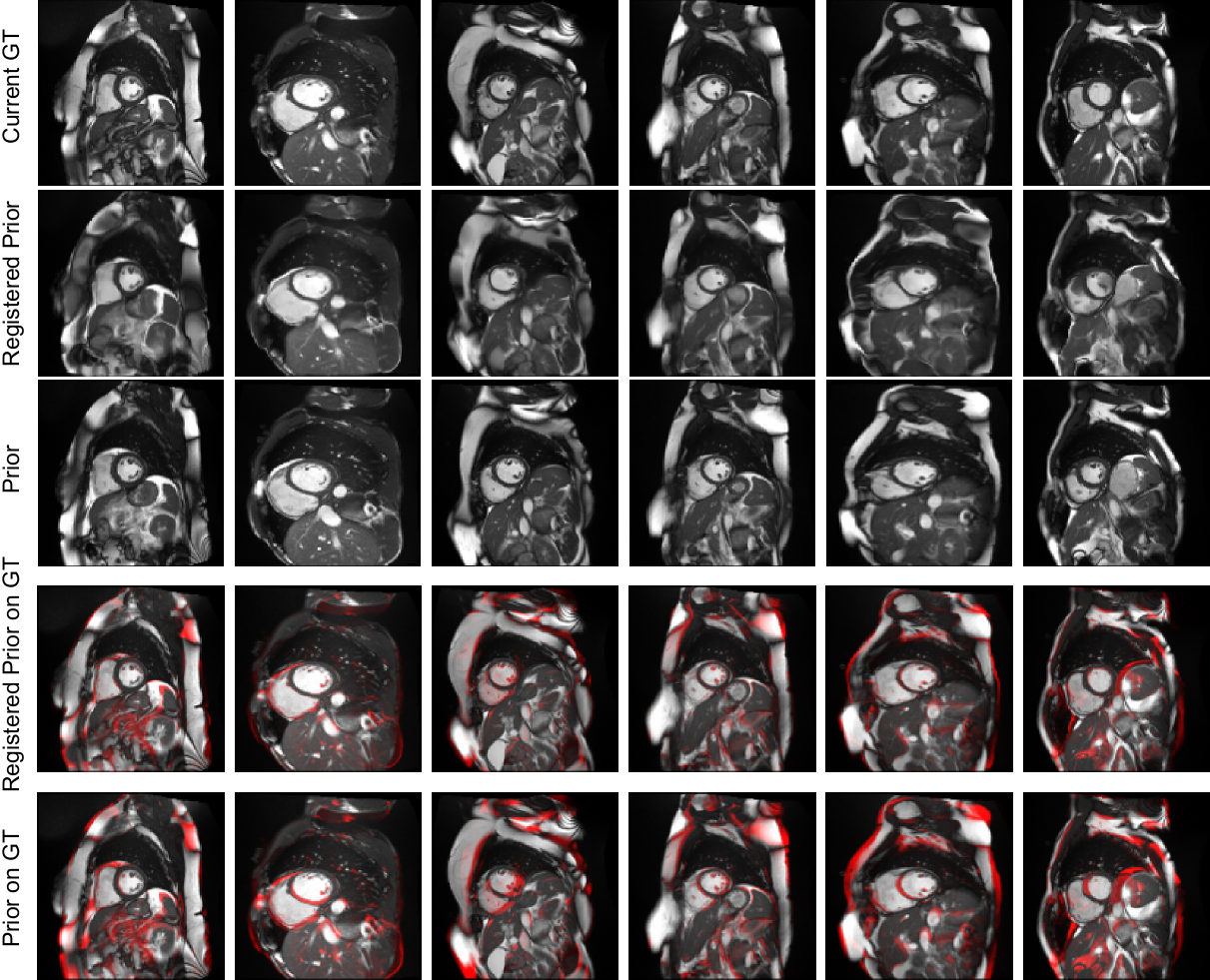}
\caption{Representative cross-visit registration audit after prior-to-current frame~0
registration. Rows show raw prior, registered prior, current reference, and raw/registered
overlays; red highlights indicate residual non-overlap.}
    \label{fig:serial_registration_audit}
\end{figure*}

Figure~\ref{fig:serial_registration_audit} shows a representative cross-visit registration audit
in which the cardiac core follows the current reference more closely after registration, while
the remaining mismatch stays concentrated around body contour, field-of-view, and other
extra-cardiac boundaries.
Taken together, these audits indicate that the main residual bottleneck in the current
cross-visit route lies in the prior-to-current registration step outside the cardiac core, rather
than in the downstream \model{} update itself. In other words, the present cross-visit
limitation is driven primarily by registration quality outside the cardiac region, not by a
failure of \model{} to track current anatomy once the cardiac core has been aligned. When
registration recovers the cardiac region well, \model{} can follow the current reference much
more closely there, which is exactly what gives the cross-visit route its ROI-facing value.

\begin{table}[t]
\centering
\caption{ROI-versus-non-ROI decomposition of prior-to-current frame~0 registration on the
62-subject dual-visit cohort. ROI denotes the current-anchor cardiac union mask (LV+RV+Myo), and
non-ROI denotes the body foreground minus ROI. Post-registration residual share denotes the
fraction of total post-registration absolute error contained within ROI versus non-ROI.}
\setlength{\tabcolsep}{4.2pt}
\renewcommand{\arraystretch}{1.10}
\begin{tabular*}{\columnwidth}{@{\extracolsep{\fill}}>{\raggedright\arraybackslash}p{31mm}rr@{}}
\toprule
Metric & ROI & non-ROI \\
\midrule
MAE (raw) & 0.1955 & 0.2823 \\
MAE (reg.) & 0.1293 & 0.2574 \\
MAE drop & 0.0662 & 0.0249 \\
SSIM (raw) & 0.6197 & 0.7345 \\
SSIM (reg.) & 0.7668 & 0.7572 \\
SSIM gain & 0.1472 & 0.0227 \\
\makecell[l]{Post-reg.\\residual share} & 0.0635 & 0.9365 \\
\bottomrule
\end{tabular*}
\label{tab:serial_registration_qc}
\end{table}

\Needspace{8\baselineskip}
\subsection{Disease-Oriented Functional Consistency}
\label{sec:disease_consistency}
We next examine downstream functional agreement across ECG-derived disease groups using both
population-level and representative-case evidence.

\subsubsection{Population-Level Consistency Across ECG-Derived Disease Categories}
\begin{figure*}[!t]
    \centering
    \includegraphics[width=\textwidth]{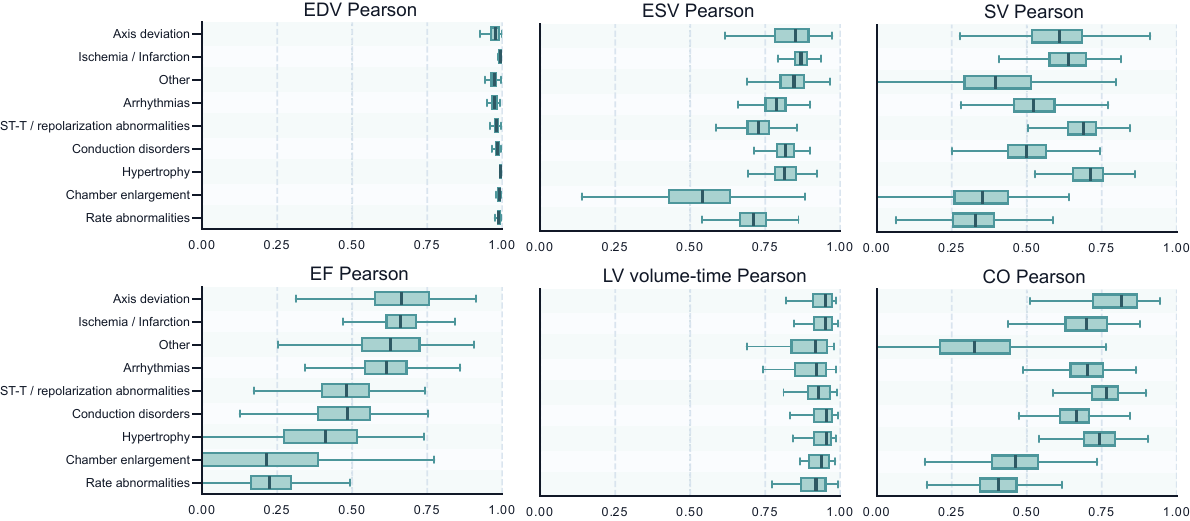}
\caption{Population-level bootstrap distributions of the selected downstream functional readouts
across ECG-derived disease categories on the shared evaluable benchmark.}
    \label{fig:disease}
\end{figure*}

Diagnosis-group labels are derived from the matched ECG metadata, and Fig.~\ref{fig:disease}
reports bootstrap distributions of downstream functional agreement across the nine retained
disease categories on the shared evaluable benchmark. Across these groups, \model{} maintains
stable functional consistency relative to the current benchmark comparator family, without any
disease category showing a qualitative reversal of the main-table pattern. The same
function-oriented stability also remains visible across heart-rate tertiles, sex strata, age
bins, and ECG-normal versus ECG-abnormal groups. Across these clinical strata, \model{} stays
within a similar local band for the main function readouts, with ESV MAE ranging from 13.58 to
19.33, EF MAE from 0.106 to 0.139, and SV MAE from 14.40 to 19.69, as summarised in
\SuppSecSubgroupSummary{} and \SuppTabSubgroupSummary{}.

\subsubsection{Representative Disease Cases and Volume Trajectories}
\begin{figure}[!t]
    \centering
    \includegraphics[width=0.96\columnwidth]{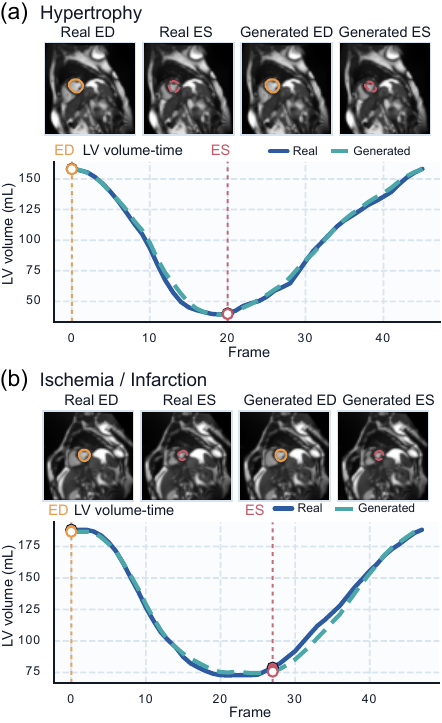}
\caption{Representative disease cases showing ED/ES geometry and LV volume--time curves for
Hypertrophy and Ischaemia / Infarction.}
    \label{fig:consistency_case}
\end{figure}

Case-level disease preservation is further illustrated with paired ED/ES geometry and LV
volume--time trajectories in Fig.~\ref{fig:consistency_case}.

Figure~\ref{fig:consistency_case} shows two representative disease cases: Hypertrophy and
Ischaemia / Infarction. The top row compares end-diastolic and end-systolic geometry in the
mid-ventricular short-axis view. The bottom row compares the corresponding real and generated
left-ventricular volume--time curves. In both cases, the generated cine preserves the case-level
geometry together with the aligned volume-curve timing pattern. The intended reading is
qualitative: these examples show interpretable case-level geometry and systolic timing rather
than providing standalone disease-specific validation.
In the hypertrophy exemplar, the generated cine preserves the case-specific cavity--myocardium
configuration and contraction--relaxation profile, whereas in the ischaemia/infarction exemplar
it retains a broader cavity appearance and more conservative systolic excursion. These
representative cases complement the cohort-level disease-family analysis and provide qualitative
case-level support that \model{} does not simply collapse to a generic average motion pattern,
although broader disease-specific validation remains necessary.

\Needspace{5\baselineskip}
\subsection{Failure Boundary}
We audit the failure boundary by first defining a hard-case pool on the same
function-and-anatomy intersection from the shared evaluable benchmark. This pool includes any
invalid outputs together with the union of the EF-error worst decile and the anatomy worst
decile, after which the resulting hard cases are manually reviewed and assigned qualitative
failure modes. Because no invalid output occurs on this intersection, the audit focuses on
degradation patterns among hard but still evaluable cases.

This audit yields 69 hard cases in total and no invalid outputs on the function+anatomy
intersection. Among these 69 cases, 43 show anatomy collapse, 22 show motion under-contraction,
3 show reference-anchor mismatch, and 1 falls into a residual other-high-error bucket,
corresponding to 62.3\%, 31.9\%, 4.3\%, and 1.4\%, respectively. Operationally, a case enters
this pool whenever it falls into the worst EF-error decile, the worst anatomy decile, or both,
after which the pooled cases are manually reviewed and assigned one dominant qualitative failure
mode. Anatomy collapse denotes cases where chamber or myocardial boundaries break down visibly
under automatic evaluation, while motion under-contraction denotes sequences that preserve the
gross chamber layout but underestimate systolic excursion or cavity shrinkage across the cardiac
cycle. The small reference-anchor mismatch group remains anatomically plausible but is
phase-shifted relative to the anchor used for evaluation, with one canonical exemplar for each
dominant failure mode shown in \SuppSecFailureBoundary{} and \SuppFigFailureBoundary{}. In
practice, the main residual motion weakness is under-contraction at systolic boundaries rather
than arbitrary temporal instability: when the model misses motion, it most often underestimates
cavity shrinkage or systolic excursion while otherwise retaining a broadly cardiac geometry. The
hard-case union has a higher ECG-abnormal fraction than the known-status cohort baseline (85.5\%
versus 78.3\%), but this difference is not significant under Fisher's exact test ($p=0.201$), so
this enrichment is descriptive only. The main failure boundary of the current method is
preserving stable contraction when anatomy collapses or systolic excursion is underestimated.

\Needspace{5\baselineskip}
\subsection{Cross-Modal Physiological Fusion}
Finally, we perform an ablation study to disentangle ECG-driven dynamics and anatomical anchoring.

\begin{table*}[!t]
\centering
\caption{Ablation study of \model{} on the UK Biobank dataset. Upper: module-wise ablations.
Lower: segmentation-consistency-loss ablations from a separate internal sweep, intended for
within-block trend reading rather than direct protocol-identical comparison to the reported
main-table COF result. `No max field' removes the maximum-displacement-field regularisation used
in this sweep.}
\setlength{\tabcolsep}{2.2pt}
\renewcommand{\arraystretch}{1.03}
\begin{tabular}{@{}lccccccccccccc@{}}
\toprule
\multirow{2}{*}{Setting} &
\multicolumn{3}{c}{\textbf{Modules}} &
\multicolumn{6}{c}{\textbf{Image metrics}} &
\multicolumn{3}{c}{\textbf{Seg.\ Dice / IoU}} \\
& REA & ECG & TOPPR &
SSIM$\!\uparrow$ &
PSNR$\!\uparrow$ &
FID$\!\downarrow$ &
FVD$\!\downarrow$ &
M\text{-}Corr.$\uparrow$ &
M\text{-}SSIM$\uparrow$ &
LV &
RV &
Myo \\
\midrule
\multicolumn{13}{l}{Module ablations} \\
COF module setting &
\checkmark & \checkmark & \checkmark &
\textbf{0.986} & \textbf{29.21} & \textbf{7.98} & \textbf{18.58} & \textbf{0.548} & \textbf{0.905} &
\multicolumn{3}{c}{N/A} \\
w/o REA &
$\times$ & \checkmark & \checkmark &
0.979 & 26.79 & 7.53 & 18.91 & 0.226 & 0.865 &
\multicolumn{3}{c}{N/A} \\
w/o Temp.\ ECG &
\checkmark & $\times$ & \checkmark &
0.978 & 26.76 & 14.50 & 25.24 & 0.512 & 0.896 &
\multicolumn{3}{c}{N/A} \\
w/o TOPPR &
\checkmark & \checkmark & $\times$ &
0.715 & 15.26 & 503.87 & 637.26 & 0.072 & 0.285 &
\multicolumn{3}{c}{N/A} \\
\midrule
\multicolumn{13}{l}{Seg.\ consistency loss (separate COF sweep; within-block trend only)} \\

No max field &
\checkmark & \checkmark & \checkmark &
0.970 & 25.16 & 11.78 & 26.40 & 0.456 & 0.895 &
0.87 / 0.78 & 0.67 / 0.51 & 0.81 / 0.68 \\

$\lambda_{\text{seg}}=0.5$ &
\checkmark & \checkmark & \checkmark &
0.982 & 27.94 & 7.34 & 17.74 & 0.533 & 0.901 &
0.88 / 0.80 & 0.71 / 0.58 & 0.84 / 0.74 \\

\textbf{$\lambda_{\text{seg}}=1.0$} &
\textbf{\checkmark} & \textbf{\checkmark} & \textbf{\checkmark} &
\textbf{0.984} & \textbf{28.46} & \textbf{6.39} & \textbf{17.60} &
\textbf{0.474} & \textbf{0.894} &
\textbf{0.89 / 0.80} & \textbf{0.74 / 0.61} & \textbf{0.85 / 0.74} \\

$\lambda_{\text{seg}}=3.0$ &
\checkmark & \checkmark & \checkmark &
0.975 & 26.04 & 12.28 & 28.43 &
0.498 & 0.901 &
0.85 / 0.76 & 0.64 / 0.50 & 0.83 / 0.71 \\
\bottomrule
\end{tabular}
\label{tab:ablation_all}
\end{table*}

In Table~\ref{tab:ablation_all}, ``REA'' denotes the reference-anatomy anchoring branch, ``ECG''
denotes temporal ECG conditioning, and ``TOPPR'' denotes the Stage~1 pairwise registration
teacher introduced above. Removing REA degrades spatial consistency and organ-shape
preservation, as reflected by drops in SSIM. Incorrectly routing the ECG embedding to the
spatial branch weakens temporal dynamics, increasing FVD and reducing motion correlation.
Removing the \registration{} causes the strongest collapse across all metrics, showing that
registration-guided flow supervision is essential for anatomy-aware phase evolution and
boundary-consistent contraction readout. The lower segmentation-consistency block is a separate
internal sweep, so its comparisons are interpreted within that block and are not
protocol-identical to the reported main-table COF result. A direct same-visit topology audit of
this Stage~1 teacher is summarised in \SuppSecTopologyAudit{} and \SuppTabTopologyAudit{},
showing near-zero folding together with high class-wise connected-component preservation on a
fixed 64-case subset.

We further vary the segmentation-consistency weight $\lambda_{\text{seg}}$ and remove the
maximum-displacement-field regularisation used in this sweep. Without segmentation consistency,
the model becomes anatomically unstable, with pronounced drops in Dice and IoU. Moderate weights
$(\lambda_{\text{seg}} \in [0.5, 1.0])$ provide the best trade-off between image quality, motion
coherence, and segmentation accuracy, whereas larger weights yield only diminishing returns.
These trends indicate that segmentation consistency mainly acts as a structural regulariser.

\section{DISCUSSION AND CONCLUSION}
\label{sec:discussion}

Taken together, this study supports ECG-conditioned cine generation from a patient-specific
anatomical anchor as a promising same-visit route for anatomy-faithful cardiac cine recovery and
downstream functional readout. The anchor-shift and 62-subject serial analyses further extend
this picture to controlled missing-target-phase proxy support and exploratory cross-visit ROI
support.

In this study, we presented \model{}, an ECG-conditioned framework that combines
patient-specific MRI and current ECG for subject-specific 4D cardiac cine generation on a
controlled same-visit benchmark. On the shared evaluable benchmark, \model{} achieves strong
image-level fidelity while preserving chamber volumetry, ejection-fraction readout, and
phase-aware functional agreement under an independent automatic evaluator. Slice-wise,
resolution-wise, disease-category, failure-boundary, and ablation analyses further indicate that
these gains depend on patient-specific MRI together with ECG-conditioned dynamics.

The current study is limited to paired in-cohort ECG-conditioned cine generation from
patient-specific MRI and to downstream assessment under a subject-disjoint automatic evaluator.
Broader manual, external, and cross-visit validation, together with longer ECG-context modeling,
remain important next steps toward clinically deployable anatomy-aware dynamic cardiac analysis.

\FloatBarrier
\phantomsection\label{bib:start}
\bibliographystyle{model2-names}
\bibliography{reference}

\end{document}